\documentclass[conference]{IEEEtran}
\IEEEoverridecommandlockouts

\usepackage{cite}
\usepackage{amsmath,amssymb,amsfonts}
\usepackage{algorithmic}
\usepackage{graphicx}
\usepackage{textcomp}
\usepackage{xcolor}

\usepackage{booktabs}
\usepackage{caption}
\usepackage{enumitem}
\usepackage{float}
\usepackage{graphicx}
\usepackage{hyperref}
\usepackage{makecell}
\usepackage{multirow}
\usepackage[table]{xcolor}
\usepackage{wrapfig}
\usepackage{xcolor}
\usepackage{pifont}
\usepackage{times}
\usepackage{tcolorbox}
\usepackage{latexsym}
\definecolor{cvprblue}{rgb}{0.21,0.49,0.74}
\usepackage{titlesec}

\usepackage{tikz}
\usepackage{xcolor}
\tcbuselibrary{skins,breakable,listings}
\def\BibTeX{{\rm B\kern-.05em{\sc i\kern-.025em b}\kern-.08em
    T\kern-.1667em\lower.7ex\hbox{E}\kern-.125emX}}

\newcommand{\bestavg}[1]{\cellcolor{orange!22}\textbf{#1}} \newcommand{\cotgain}[1]{\cellcolor{green!14}#1}

\begin{document}

\title{CinematicVQA: Benchmarking Film-Grammar Reasoning in Large Vision-Language Models
}

\author{%
\IEEEauthorblockN{1\textsuperscript{st} Shuo Xing$^*$}\thanks{$^*$\,This work was done while the author interned at Google.}
\IEEEauthorblockA{\textit{Department of Computer Science \& Engineering} \\
\textit{Texas A\&M University}\\
College Station, United States \\
shuoxing@tamu.edu}
\and
\IEEEauthorblockN{2\textsuperscript{nd} Pooja Verlani}
\IEEEauthorblockA{\textit{Google Inc.} \\
Mountain View, United States \\
poojatandon@google.com}

\IEEEauthorblockN{3\textsuperscript{rd} Balu Adsumilli}
\IEEEauthorblockA{\textit{Google Inc.} \\
Mountain View, United States \\
badsumilli@google.com}
\and
\IEEEauthorblockN{4\textsuperscript{th} Zhengzhong Tu}
\IEEEauthorblockA{\textit{Department of Computer Science \& Engineering} \\
\textit{Texas A\&M University}\\
College Station, United States \\
tzz@tamu.edu}
}

\author{
\IEEEauthorblockN{
Shuo Xing$^1$$^*$\thanks{$^*$\,This work was done while the author interned at Google.},
Pooja Verlani$^2$,
Balu Adsumilli$^2$,
Zhengzhong Tu$^1$
}

\vspace{1em}

\IEEEauthorblockA{
$^1$Texas A\&M University, College Station, USA \quad \quad $^2$Google Inc., Mountain View, USA \\
}
}

\definecolor{myred}{HTML}{E14169} 
\definecolor{royalblue}{HTML}{4169E1}

\newcommand{\revision}[1]{\textcolor{purple}{#1}}
\newcommand{\cqat}{{\textsc{CinematicVQA-train}}}
\newcommand{\cqae}{{\textsc{CinematicVQA-eval}}}
\newcommand{\cqa}{{CinematicVQA}}

\newcommand{\cmark}{\ding{51}} 
\newcommand{\xmark}{\ding{55}} 

\newtcolorbox{boxK}[2][]{
    sharpish corners, 
    boxrule = 0pt,
    toprule = 4.5pt, 
    enhanced,
    fuzzy shadow = {0pt}{-2pt}{-0.5pt}{0.5pt}{black!35}, 
    fontupper = \ttfamily\small, 
    boxsep = 5pt, 
    left = 5pt, 
    right = 5pt, 
    top = 5pt, 
    bottom = 5pt, 
    #1                       
}

\newtcolorbox{takeawaybox}[2][]{
    enhanced,
    boxsep = 2pt, 
    left = 2pt, 
    right = 2pt, 
    top = 2pt, 
    bottom = 2pt, 
    #1                       
} 

\maketitle

\begin{abstract}
Cinematography, the craft of visual storytelling through framing, lighting, and camera operation, fundamentally shapes how audiences perceive and emotionally engage with video content. While Large Vision Language Models (LVLMs) have made remarkable progress in video question answering, existing benchmarks primarily focus on identifying low-level techniques rather than understanding their storytelling impact. To address this, we introduce CinematicVQA, the first-of-its-kind benchmark for cinematic video understanding that goes beyond technique recognition to evaluate film-grammar reasoning, utilizing our introduced Cinematic Scene Graph (CSG), a structured representation that links filming techniques to their perceptual effects and narrative functions. Through comprehensive evaluation of state-of-the-art LVLMs, we reveal a striking semantic gap: models consistently perform higher on describing visual presentations than on identifying the underlying techniques. Surprisingly, Chain-of-Thought prompting fails to provide consistent gains and degrades performance for most models, suggesting that current LVLMs lack sufficient cinematic domain knowledge to benefit from step-by-step reasoning. Fine-tuning on \cqat{} yields consistent improvements, particularly for narrative function and multi-hop reasoning. Overall, \cqa{} serves both as a rigorous benchmark for cinematic evaluation in LVLMs and as a practical dataset for training more film-aware video models.
\end{abstract}

\begin{IEEEkeywords}
Large Vision Language Models, Video Understanding, Cinematography, Benchmarking
\end{IEEEkeywords}

\section{Introduction}

Cinematography is a highly structured visual language: video makers deliberately manipulate composition, lighting, and camera movement to steer attention, shape mood, and communicate narrative intent. LVLMs have shown strong performance on general-purpose video question answering (VQA), as well show promising progress on cinematography-oriented tasks~\cite{huang2020movienet,savardi2023cinescale2,lin2025towards,wang2025cinetechbench,liu2025shotbench,wu2025refineshot}. However, existing benchmarks still primarily focus on technique identification, i.e., what compositional, lighting, or camera-operation technique appears in a shot, rather than how a shot is constructed for storytelling: the perceptual effects these choices create and the narrative functions they support. However, this overlooks a critical reality, that the true value of these techniques lies in their ability to serve the narration, a requisite for creating engaging and aesthetically driven content.

To address this gap, we introduce \cqa{}, the first benchmark designed to evaluate and advance comprehensive cinematic reasoning in LVLMs. \cqa{} targets cinematography-aware understanding along three core dimensions, including composition, lighting, and camera operation. Notably, it goes beyond treating cinematographic techniques as isolated labels, instead probing the structure of cinematic reasoning by: (i) identifying technique-relevant visual evidence in the video clip; (ii) connecting techniques to the perceptual effects they induce; and (iii) inferring plausible narrative intent and storytelling function conditioned on those effects.
\begin{figure*}[t]
    \centering
    \includegraphics[width=0.9\linewidth]{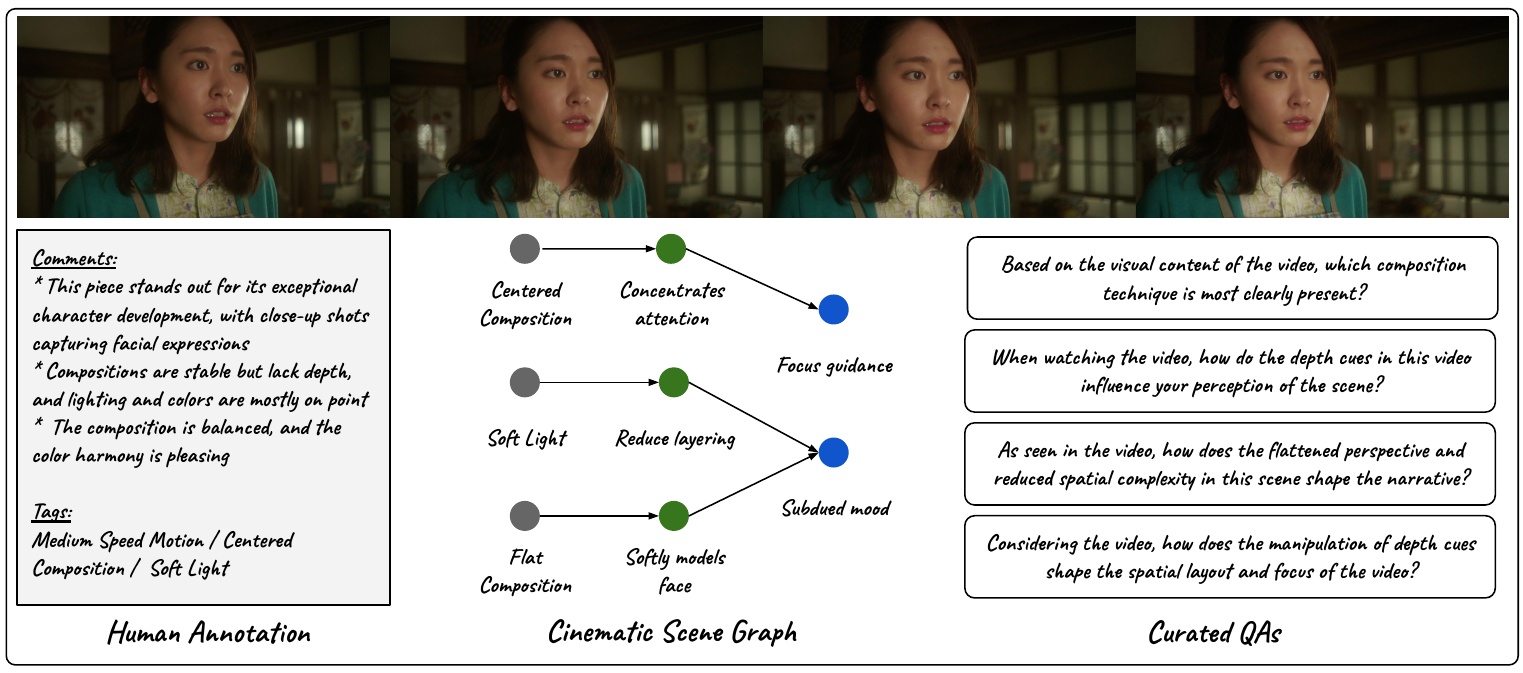}
    \caption{{Overview of the proposed \cqa{}. A VQA benchmark designed to comprehensively evaluate the cinematic understanding of LVLMs from the recognition of filming techniques to their perceptual and narrative consequences.}
    }
    \label{fig:teaser}
    \vspace{-0.17in}
\end{figure*}
However, building such a benchmark is challenging as the cinematic meaning is often implicit, subjective, and context-dependent. To anchor our supervision in human interpretation, we build \cqa{} upon the Video Aesthetic Database (VADB)~\cite{qiao2025vadb}, utilizing its expert annotations. Specifically, for each video segment, we construct a Cinematic Scene Graph (CSG) by parsing and normalizing expert commentary into structured elements. This representation explicitly links cinematographic techniques (the mechanism) with their corresponding perceptual presentations and narrative impacts (the outcome). In summary, our contributions are summarized as follows:
\begin{itemize}[leftmargin=*,nosep]
    \item We present \cqa{}, a first-of-its-kind benchmark for cinematic video understanding that goes beyond technique recognition to evaluate film-grammar reasoning—how models identify and interpret composition, lighting, and camera operation, and connect these choices to their perceptual effects and storytelling implications.


    \item We introduce a cinema-aware data curation pipeline centered on the Cinematic Scene Graph (CSG), which transforms expert-level cinematic annotations into structured and interpretable VQA pairs. CSGs decompose the cinematography of an input video clip into nodes spanning filming techniques, visual presentation, and narrative effects, and link them with typed edges that formalize their dependencies, e.g., $technique \rightarrow presentation \rightarrow narrative$.

    \item We conduct a comprehensive evaluation of representative LVLMs on \cqae{} and find that supervised fine-tuning (SFT) on \cqat{} yields consistent gains in cinematic reasoning, highlighting \cqa{} as both a rigorous benchmark for evaluation and a practical resource for training film-aware models.
\end{itemize}

\section{Related Work}

\paragraph{Large Vision Language Models} 

Recent advancements in LVLMs~\cite{liu2024llava,gpt4omini,gemini3,bai2025qwen2.5vl,Qwen3-VL,wang2025internvl3_5} have significantly bridged the gap between textual and visual understanding built upon the success of the Large Language Models (LLMs)~\cite{devlin2018bert, brown2020gpt3,team2023gemini,touvron2023llama2,qwen2,yang2025qwen3}. Typically, the foundational paradigm of LVLMs couples a vision encoder (e.g., CLIP~\cite{radford2021clip}) with an LLM backbone via a linear projector. Specifically, input images are first encoded into image tokens, projected into the language embedding space, and concatenated with text embeddings and fed into the LLM for autoregressive generation, thereby unlocking unprecedented cross-modal applications in real-world scenarios~\cite{moor2023med, openemma}.

\paragraph{Cinematic Understanding Benchmarks}

Existing benchmarks for cinematic understanding in LVLMs primarily focus on the recognition of low-level cinematography techniques~\cite{lin2025towards,wang2025cinetechbench,liu2025shotbench,wu2025refineshot}. Specifically, CameraBench~\cite{lin2025towards} provides a fine-grained evaluation of camera motion, while CineTechBench~\cite{wang2025cinetechbench} covers a diverse taxonomy of general cinematographic techniques. Based on this, ShotBench~\cite{liu2025shotbench} introduces a broader set of evaluation dimensions, which is further refined by RefineShot~\cite{wu2025refineshot}. However, these works lack evaluation metrics for higher-level understanding, such as analyzing how specific techniques influence visual aesthetics and contribute to storytelling. As depicted in Table \ref{tab:comp}, our proposed \cqa{} addresses this limitation by introducing the first multi-level cinematic evaluation protocol, extending beyond atomic recognition.

\begin{table}[htbp]
  \footnotesize
  \setlength{\tabcolsep}{4pt}

  \begin{center}
    \begin{tabular}{lcccc}
      \toprule

      Dataset
      & \makecell[c]{Low-level\\Technique}
      & \makecell[c]{Mid-level\\Visual Effect}
      & \makecell[c]{High-level\\Narrative Function}
      \\

      \midrule

      CameraBench~\cite{lin2025towards}
      & \cellcolor{green!12}\cmark
      & \cellcolor{green!12}\cmark
      & \cellcolor{red!12}\xmark
      \\

      CineTechBench~\cite{wang2025cinetechbench}
      & \cellcolor{green!12}\cmark
      & \cellcolor{red!12}\xmark
      & \cellcolor{red!12}\xmark
      \\

      ShotBench~\cite{liu2025shotbench}
      & \cellcolor{green!12}\cmark
      & \cellcolor{red!12}\xmark
      & \cellcolor{red!12}\xmark
      \\

      RefineShot~\cite{wu2025refineshot}
      & \cellcolor{green!12}\cmark
      & \cellcolor{red!12}\xmark
      & \cellcolor{red!12}\xmark
      \\

      \textbf{\cqa{}}
      & \cellcolor{green!12}\cmark
      & \cellcolor{green!12}\cmark
      & \cellcolor{green!12}\cmark
      \\

      \bottomrule
    \end{tabular}
  \end{center}
  \caption{Comparison between \cqa{} and existing LVLM VQA benchmarks.}
  \label{tab:comp}
  \vspace{-0.25in}
\end{table}

\section{CinematicVQA}

In this section, we present CinematicVQA, the first-of-its-kind VQA benchmark designed to comprehensively evaluate the cinematic understanding of LVLMs from the recognition of filming techniques to their perceptual and narrative consequences. First, we introduce the Cinematic Scene Graph (CSG), a structured representation that explicitly models the causal and relational links between filming techniques, visual presentation, and narrative impact. We then curate CinematicVQA by building upon the public VADB dataset, leveraging its human-written aesthetic comments to construct CSGs at scale and generate the \cqat{} (training) and \cqae{} (evaluation). 


\subsection{Cinematic Scene Graph}

Given a video clip $v$, we define its \emph{Cinematic Scene Graph} (CSG) as a directed, typed graph
$\mathcal{G}_v = (\mathcal{V}_v, \mathcal{E}_v)$.
The node set $\mathcal{V}_v$ is partitioned into three semantic types:
\begin{itemize}[leftmargin=*, nosep]
    \item \textbf{Technique nodes} $\mathcal{T}_v$, where each $t\in\mathcal{T}_v$ represents a canonical filming technique restricted to one of three categories: \emph{Composition}, \emph{Lighting}, or \emph{Camera Operation}; each technique node is accompanied by \emph{evidence} grounded in the input text.

    \item \textbf{Presentation-effect nodes} $\mathcal{P}_v$, where each $p\in\mathcal{P}_v$ describes how a technique modulates visual perception or presentation (e.g., emphasis, saliency, stability, pacing, atmosphere), without describing objects, characters, or events.

    \item \textbf{Narrative-function nodes} $\mathcal{N}_v$, where each $n\in\mathcal{N}_v$ captures an abstract narrative or emotional function (e.g., focus guidance, tension building, emotional framing).
\end{itemize}

For the distinct types of nodes, we have: $\mathcal{V}_v = \mathcal{T}_v \sqcup \mathcal{P}_v \sqcup \mathcal{N}_v$.
Edges $\mathcal{E}_v$ are directed and typed: each edge is a triple $(u, r, w)$ connecting two nodes $u,w\in\mathcal{V}_v$ with relation type
$r \in \{\texttt{enhances}, \texttt{reinforces}, \texttt{causes}, \texttt{limits},$ $ \texttt{supports}, \texttt{results\_in}\}$.

We impose structural constraints to encode a directed causal reasoning chain: (a) every technique node must connect to at least one presentation-effect node, and (b) every presentation-effect node must connect to at least one narrative-function node, enabling paths of the form $t \rightarrow p \rightarrow n$ that represent \emph{technique $\rightarrow$ perception $\rightarrow$ narrative}.

\subsection{Data Curation}

\begin{figure}[t]
    \centering
    \includegraphics[width=0.85\linewidth]{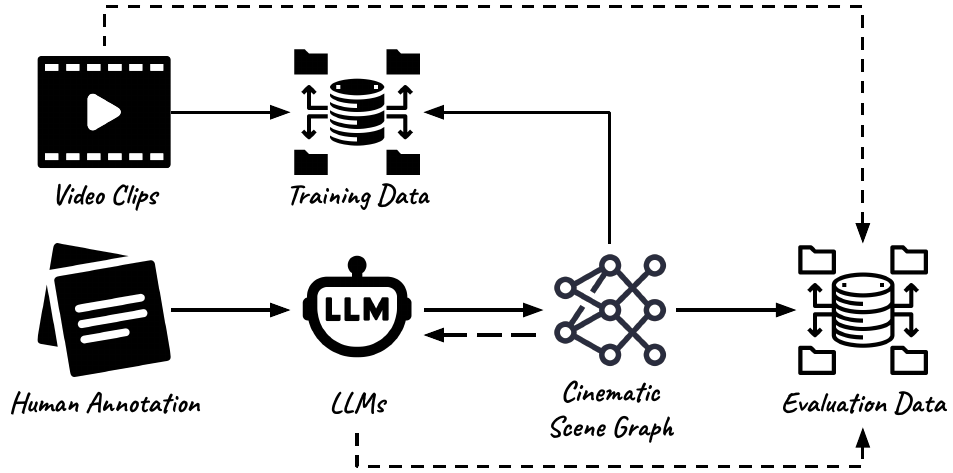}
    \caption{{Illustration of the data curation pipeline.}
    }
    \label{fig:data-gen}
    \vspace{-0.17in}
\end{figure}

Given the dense semantic signals required to construct a CSG, we build upon the public Video Aesthetic Description Benchmark (VADB) by leveraging its human-written comments. Specifically, We construct $\mathcal{G}_v$ for each video $v$, following the pipeline illustrated in Figure \ref{fig:data-gen}.

\begin{itemize}[leftmargin=*, nosep]
    \item \textbf{Input formatting.} For each $v$, we collect \emph{caption tags} and \emph{free-form captions}; captions are preserved with explicit sentence indices to support evidence attribution.

    \item \textbf{Constrained extraction.} We prompt an LLM (GPT-4o mini~\cite{gpt4omini}) to extract film-grammar attributes across three distinct categories: Composition, Lighting, and Camera Operation. The model is tasked with generating a structured graph comprising: (i) technique nodes anchored by textual evidence, (ii) presentation-effect nodes detailing perceptual consequences, (iii) narrative-function nodes capturing abstract intent, and (iv) directed edges representing the relationships between these components.

    \item \textbf{VQA Curation.} We sample one technique-to–narrative-function path per category (composition, lighting, and camera operation). For each sampled path, we instantiate the template corresponding to the target reasoning hop to generate the question, and we use the subsequent node along the path as the ground-truth answer, thereby producing the VQA pairs. We then split these pairs into \cqat{} and \cqae{} sets with an approximately 9:1 ratio, yielding 42,748 training VQA pairs and 4,596 evaluation VQA pairs. We further use LLMs to refine question-answer quality and generate plausible multiple-choice distractors. Consequently, for each input video and each technique category, we construct four QA pairs that probe complementary dimensions of cinematic analysis:
    \begin{itemize}[leftmargin=*, nosep]
        \item \textbf{Technique Recognition.} Identifies the specific cinematic technique present in the clip.
        \item \textbf{Visual Presentation.} Describes how the technique manifests on screen, i.e., its observable visual realization and effect on the clip’s appearance.
        \item \textbf{Narrative Function.} Interprets the semantic intent or storytelling role of the resulting visual presentation (e.g., conveying tension or unease).
        \item \textbf{Multi-Hop Reasoning.} Requires chaining across levels by first recognizing the technique and then explaining how it produces a particular visual presentation that supports the narrative function.
    \end{itemize}

\end{itemize}

\subsection{Human Annotations on Data Quality}

To evaluate the quality of the both the generated Cinematic Scene Graphs (CSGs) and the CinematicVQA test set, three human experts (junior and senior graduate students) independently scored a subset of the data (100 samples each) using the rubrics provided in Figure \ref{fig:rubric}.

\begin{figure}[htbp]
    \centering
    \begin{boxK}[colback=royalblue!5!white, colframe=royalblue!75!black, fontupper=\ttfamily\tiny]
    
    \textbf{\#\# Task 1: Cinematic Scene Graph Quality:} \\
        \begin{itemize}[leftmargin=0.15in]
            \item 1 - Critical Failure: Severe hallucinations; fabricated nodes or edges; misses primary subjects entirely.
            \item 2 - Poor: Captures basic objects but fails on crucial relationships or attributes; sparse mapping.
            \item 3 - Acceptable: Mostly accurate but lacks density; omits secondary edge details; no major hallucinations.
            \item 4 - Good: Accurately maps primary and secondary objects, attributes, and relationships; minor omissions only.
            \item 5 - Excellent: Dense, perfectly grounded, and highly structured; zero hallucinations; captures complex spatial and semantic nuances.
        \end{itemize}
       \ \\
    \textbf{\#\# Task 2: CinematicVQA MCQ Distractor Quality:} \\
        \begin{itemize}[leftmargin=0.15in]
            \item 1 - Critical Failure: Distractors are trivial, nonsensical, logically disjointed from the video sequence, or accidentally correct.
            \item 2 - Poor: Easily guessable using static frame bias or language priors; fails to test temporal reasoning, action recognition over time, or dialogue context.
            \item 3 - Acceptable: Plausible within the scene but can be eliminated by observing a single keyframe or relying on shallow, short-term visual actions.
            \item 4 - Good: Strong distractors requiring the model to track entities across shot transitions, understand temporal ordering, or resolve causal chains to identify the correct answer.
            \item 5 - Excellent: Expertly crafted hard negatives; specifically target cinematic failure modes requiring deep reasoning.
        \end{itemize}
    \end{boxK}

    \caption{Human evaluation rubrics for assessing Cinematic Scene Graph (CSG) quality and \cqae{} multiple-choice distractor quality. Experts scored each sample on a 1--5 scale, where higher scores indicate better annotation quality.}
    \label{fig:rubric}
\end{figure}

The generated CSGs achieved an average score of \textbf{\textit{4.36}}, while the \cqae{} multiple-choice distractors achieved an average score of \textbf{\textit{4.07}}. These human evaluation results provide direct evidence for the quality of our constructed dataset. Specifically, the high CSG score indicates that the generated scene graphs are well-grounded and structurally coherent, while the strong distractor score suggests that the answer choices are non-trivial and effective for evaluating cinematic understanding rather than shallow pattern matching.

\begin{table*}[htbp]
  \footnotesize
  \setlength{\tabcolsep}{4.5pt}
  \begin{center}
  \begin{tabular}{l
  ccc    
  ccc    
  ccc    
  ccc    
  |c}
  \toprule
  & \multicolumn{3}{c}{\cellcolor{green!12}\textbf{Technique Recognition}}
  & \multicolumn{3}{c}{\cellcolor{gray!12}\textbf{Visual Presentation}}
  & \multicolumn{3}{c}{\cellcolor{blue!12}\textbf{Narrative Function}}
  & \multicolumn{3}{c}{\cellcolor{yellow!12}\textbf{Multi-Hop Reasoning}}
  &  \\
  \cmidrule(lr){2-4}\cmidrule(lr){5-7}\cmidrule(lr){8-10}\cmidrule(lr){11-13}
  \textbf{Model}
  & Comp & Light & Camera
  & Comp & Light & Camera
  & Comp & Light & Camera
  & Comp & Light & Camera
  & \textbf{Avg.} \\

  \midrule
  \multicolumn{14}{c}{\textit{Zero-Shot Prompting}} \\
  \midrule

  InternVL-3.5-4B
    & 39.16 & 45.69 & 38.12
    & 52.74 & 54.31 & 59.01
    & 47.52 & 37.60 & 37.34
    & 49.87 & 37.34 & 47.78
    & 45.54$_{\pm 0.12}$ \\

  \rowcolor{orange!6}
  InternVL-3.5-8B
    & 40.21 & 45.69 & 37.34
    & 55.87 & 54.57 & 60.05
    & 46.48 & 41.25 & 37.34
    & 53.26 & 45.43 & 46.21
    & \bestavg{46.98$_{\pm 0.10}$} \\

  Qwen2.5-VL-3B-it
    & 39.86 & 37.77 & 34.64
    & 52.39 & 47.43 & 52.91
    & 51.61 & 41.16 & 33.85
    & 48.47 & 48.21 & 39.34
    & 43.97$_{\pm 0.16}$ \\

  Qwen2.5-VL-7B-it
    & 43.42 & 40.29 & 36.63
    & 56.22 & 47.86 & 54.65
    & 46.82 & 42.38 & 36.11
    & 45.25 & 46.82 & 40.81
    & 44.77$_{\pm 0.09}$ \\

  Qwen3-VL-4B-it
    & 30.23 & 35.45 & 33.88
    & 59.47 & 50.07 & 60.25
    & 43.28 & 34.93 & 34.14
    & 51.38 & 40.15 & 41.19
    & 42.87$_{\pm 0.07}$ \\

  Qwen3-VL-8B-it
    & 29.12 & 35.39 & 31.21
    & 60.45 & 46.62 & 56.54
    & 41.66 & 34.87 & 33.30
    & 51.05 & 44.79 & 48.97
    & 42.83$_{\pm 0.17}$ \\

  Qwen3-VL-32B-it
    & 28.72 & 34.19 & 30.41
    & 59.95 & 45.82 & 55.44
    & 41.11 & 34.07 & 32.25
    & 50.60 & 43.99 & 47.82
    & 42.03$_{\pm 0.11}$ \\

  \midrule
  \multicolumn{14}{c}{\textit{Chain-of-Thought Prompting}} \\
  \midrule

  InternVL-3.5-4B
    & 40.15 & 42.24 & 35.97
    & 50.07 & 51.64 & 59.99
    & 40.67 & 34.14 & 32.32
    & 47.98 & 35.71 & 43.28
    & 42.85$_{\pm 0.15}$ \\

  \rowcolor{orange!6}
  InternVL-3.5-8B
    & 40.39 & 41.43 & 38.56
    & 52.66 & 51.09 & 62.32
    & 41.96 & 40.39 & 33.34
    & 52.40 & 44.05 & 45.09
    & \bestavg{45.31$_{\pm 0.11}$} \\

  Qwen2.5-VL-3B-it
    & 33.24 & 27.50 & 33.76
    & 50.21 & 41.07 & 41.33
    & 51.25 & 34.81 & 35.85
    & 41.59 & 36.11 & 26.71
    & 37.79$_{\pm 0.18}$ \\

  Qwen2.5-VL-7B-it
    & 29.43 & 30.48 & 29.17
    & 51.37 & 45.88 & 58.68
    & 39.09 & 35.18 & 32.83
    & 47.45 & 46.67 & 41.97
    & 40.68$_{\pm 0.13}$ \\

  Qwen3-VL-4B-it
    & 27.74 & 37.66 & 30.35
    & 61.16 & 55.93 & 64.29
    & 39.49 & 37.14 & 38.96
    & 49.67 & 41.05 & 43.40
    & \cotgain{43.90$_{\pm 0.08}$} \\

  Qwen3-VL-8B-it
    & 29.70 & 37.79 & 30.74
    & 53.19 & 48.49 & 59.72
    & 41.71 & 36.22 & 37.53
    & 47.97 & 41.97 & 46.41
    & 42.62$_{\pm 0.16}$ \\

  Qwen3-VL-32B-it
    & 28.50 & 35.80 & 29.90
    & 57.20 & 49.80 & 61.50
    & 42.30 & 36.80 & 36.50
    & 49.60 & 42.40 & 40.66
    & \cotgain{42.58$_{\pm 0.13}$} \\

  \bottomrule
  \end{tabular}
  \caption{Performance of state-of-the-art LVLMs on CinematicVQA across four reasoning categories: Technique Recognition, Visual Presentation, Narrative Function, and Multi-Hop Reasoning. We highlight the best overall average within each prompting setting in orange, and mark cases where chain-of-thought prompting improves over the corresponding zero-shot result in green.}
  \label{tab:eval-res}
  \end{center}
  \vspace{-0.2in}
\end{table*}

\section{Experiment}

In this section, we conduct a comprehensive evaluation of state-of-the-art LVLMs on \cqae{} under both zero-shot and Chain-of-Thought (CoT)~\cite{wei2022chain} prompting. We further perform supervised fine-tuning (SFT) on \cqat{} to assess the utility of the proposed training set and its transfer to cinematic reasoning. Overall, our experiments are structured around two primary questions: (1) \textit{To what extent can current LVLMs understand and reason about cinematic techniques and their associated visual–narrative functions in videos?} and (2) \textit{How much does \cqat{} improve LVLMs’ cinematic perception and reasoning capabilities, including technique recognition, visual effect attribution, and narrative-function inference? }

\subsection{Experimental Setup}

For evaluation, we benchmark a diverse set of state-of-the-art LVLMs, including InternVL-3.5 (4B \& 8B)~\cite{wang2025internvl3_5}, Qwen2.5-VL-Instruct (3B \& 7B)~\cite{bai2025qwen2}, and Qwen3-VL-Instruct (4B, 8B, \& 32B)~\cite{Qwen3-VL}. We evaluate all models under both zero-shot and CoT prompting on \cqae{}. To ensure consistency, each model is provided with an identical prompt template and uses a fixed maximum generation budget of 100 tokens for zero-shot and 512 tokens for CoT. Model outputs are converted into final multiple-choice predictions via a rule-based answer extractor that applies multiple regular-expression patterns to identify the selected option letter. For SFT finetuning, we apply LoRA with rank 128 and alpha 256, and train for one epoch with a per-device batch size of 1 and 2 gradient accumulation steps. We use a learning rate of 1e-6 with a linear warmup ratio of 0.05, training on 8 NVIDIA A6000 Ada GPUs. All models use BF16 mixed precision for both training and inference.

\subsection{Benchmark with \cqae{}}

Table \ref{tab:eval-res} demonstrates the performance of the evaluated model on \cqae{} across the technique categories and cinematic understanding dimensions. 

\paragraph{\textit{General Performance.}}
Overall, model performance on \cqa{} remains moderate. Notably, Technique Recognition consistently trails Visual Presentation, suggesting that the primary bottleneck is not raw visual perception, but rather a lack of grounded film-grammar knowledge.
Among the evaluated models, the InternVL-3.5-8B delivers the strongest overall average among all evaluated models under both zero-shot and CoT prompting. Notably, the closed-source models we evaluated, GPT-4o mini~\cite{gpt4omini} (40.69) and Gemini-2.5-Flash~\cite{comanici2025gemini} (41.32), perform at a similar level to the open-source models.

Surprisingly, however, Chain-of-Thought (CoT) prompting fails to deliver consistent gains and degrades accuracy for most models compared to standard zero-shot inference, highlighting that current LVLMs lack sufficient grounding in cinematic domain knowledge to effectively utilize step-by-step reasoning without drifting from the visual evidence.


\paragraph{Where LVLMs Still Struggle.}Generally, all models struggle with technique recognition tasks. We attribute this to the dual challenge of the task, which requires both fine-grained visual grounding capabilities and specialized domain knowledge of cinematic filmmaking. Crucially, the results reveal a ``semantic gap'' between perception and terminology: models consistently score 10--20\% higher on \textit{Visual Presentation} than on \textit{Technique Recognition}. For instance, under zero-shot prompting, Qwen3-VL-8B-it achieves a score of 60.45\% in describing the visual appearance of composition, yet achieves only 29.12\% in correctly identifying those specific composition techniques, suggesting that while current LVLMs can effectively perceive and describe visual patterns, they frequently lack the precise domain vocabulary to classify them.

\subsection{Does Chain-of-Thought Help on \cqa{}?}

A natural hypothesis is that explicit reasoning should help on a benchmark that asks not only \emph{what} cinematographic technique is used but also \emph{how} it shapes the viewer's experience. Yet evaluating every open-source VLM with a chain-of-thought (CoT) prompt contradicts this intuition (Table~\ref{tab:eval-res}): across seven models spanning two families and scales from 3B to 32B, CoT yields no overall mean gain, degrading the Qwen2.5 models most sharply and InternVL mildly while the Qwen3 family stays within noise. This near-zero mean reflects \emph{cancellation} rather than \emph{stability}---on Qwen3-VL-8B, CoT changes the predicted option on 34.6\% of questions, but losses and gains almost exactly offset ($472$ correct$\to$incorrect vs.\ $460$ incorrect$\to$correct; McNemar $\chi^2_1 = 0.13$, $p = 0.72$; Table~\ref{tab:flips}). CoT thus perturbs roughly one in three predictions in both directions instead of stabilizing the answers the model already gets right, indicating that current VLMs lack reliable cinematic-reasoning priors: eliciting longer textual reasoning is insufficient for cinematic understanding and can even displace correct visual judgments.

\begin{table}[htbp]
  \centering
  \small
  \begin{tabular}{l cc c}
  \toprule
  & \multicolumn{2}{c}{\textbf{CoT}} & \\
  \cmidrule(lr){2-3}
  \textbf{Zero-shot} & Correct & Incorrect & Total \\
  \midrule
  Correct   & 1{,}400 & 472     & 1{,}872 \\
  Incorrect & 460     & 2{,}162 & 2{,}622 \\
  \midrule
  Total     & 1{,}860 & 2{,}634 & 4{,}494 \\
  \bottomrule
  \end{tabular}
  \caption{Paired zero-shot vs.\ CoT outcomes for Qwen3-VL-8B on the $4{,}494$ questions answerable under both prompts ($102$ are unparseable under CoT and excluded). The off-diagonal flips nearly cancel, and a McNemar test finds no significant difference ($\chi^2_1 = 0.13$, $p = 0.72$).}
  \label{tab:flips}
  \vspace{-0.11in}
\end{table}



Inspecting the cases where CoT turns a correct zero-shot answer wrong reveals a single underlying mechanism: the model commits to a free-form description first, then selects the option that best matches its \emph{own words} rather than the video. This surfaces as five recurring patterns, summarised in Table~\ref{tab:cot-failures}.

\begin{table}[htbp]
\footnotesize
\setlength{\tabcolsep}{4pt}
\renewcommand{\arraystretch}{1.25}
\begin{tabular}{@{}>{\raggedright\arraybackslash}p{2.0cm}>{\raggedright\arraybackslash}p{5.75cm}@{}}
\toprule
\textbf{Failure mode} & \textbf{Mechanism} \\
\midrule
Description displaces evidence & The chain narrates the clip, then matches its \emph{own words} to the options instead of re-grounding on the footage. \\
Generic-distractor bias & Broad, neutral wording composes most naturally with the blandest option, which is over-selected. \\
Negation misalignment & Enumerating what is \emph{not} seen aligns literally with the foil that recycles a just-written word, over the broader correct option. \\
Adjective inertia & After an adjective (``stable'', ``smooth'') is written, later steps favour the option that reuses it, conflating distinct concepts. \\
``Trick-question'' spiral & With no option matching its description, the model loops in self-doubt and exhausts its budget without committing to a letter. \\
\bottomrule
\end{tabular}
\caption{Recurring CoT failure modes from the right-to-wrong cases; all share one cause---the model matches its own generated text rather than the video.}
\label{tab:cot-failures}
\vspace{-0.15in}
\end{table}

\subsection{Finetuning with \cqat{}}
\begin{figure}[htbp]
    \centering
    \includegraphics[width=\linewidth]{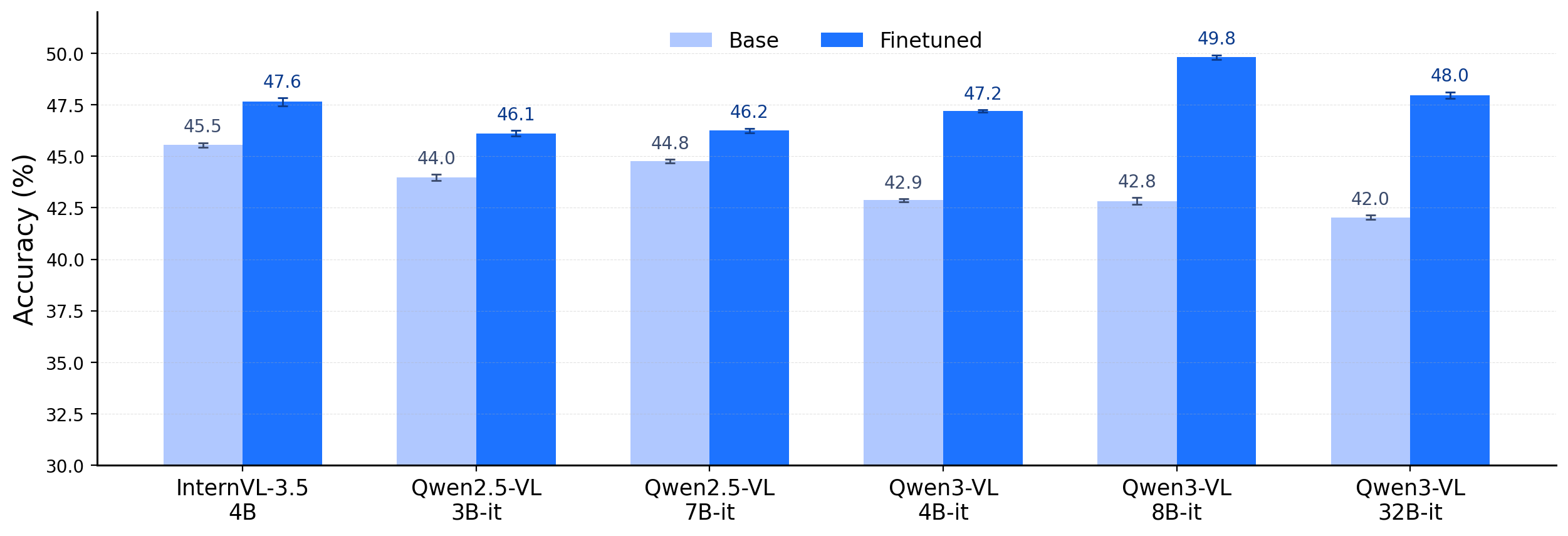}
    \caption{Comparison of base and fine-tuned model performance across different model scales. Bars show performance (accuracy) on the \cqae{}, with each model’s fine-tuned version plotted adjacent to its base version.
    }
    \label{fig:ft-model}
    \vspace{-0.1in}
\end{figure}

We perform SFT with \cqat{} on InternVL-3.5 (4B), Qwen2.5-VL-Instruct (3B \& 7B), and Qwen3-VL-Instruct (4B,
8B, \& 32B) as backbones, and evaluate on \cqae{} under zero-shot prompting. The performance of the fine-tuned models is summarized in Table \ref{tab:finetune}, and the corresponding relative improvements over their vanilla models are reported in Figure \ref{fig:ft-model}.

\begin{table*}[htbp]
  \footnotesize
  \setlength{\tabcolsep}{4.5pt}
  \begin{center}
  \begin{tabular}{l ccc ccc ccc ccc |c}
  \toprule
  & \multicolumn{3}{c}{\cellcolor{green!12}\textbf{Technique Recognition}}
  & \multicolumn{3}{c}{\cellcolor{gray!12}\textbf{Visual Presentation}}
  & \multicolumn{3}{c}{\cellcolor{blue!12}\textbf{Narrative Function}}
  & \multicolumn{3}{c}{\cellcolor{yellow!12}\textbf{Multi-Hop Reasoning}}
  & \\
  \cmidrule(lr){2-4}\cmidrule(lr){5-7}\cmidrule(lr){8-10}\cmidrule(lr){11-13}
  \textbf{Model ($\Delta$)}
  & Comp & Light & Camera
  & Comp & Light & Camera
  & Comp & Light & Camera
  & Comp & Light & Camera
  & \textbf{Avg.} \\

  \midrule
  
  InternVL-3.5-4B (FT)
  & {\color{red!60!black} -1.50} & {\color{red!60!black} -2.20} & {\color{red!60!black} -0.80}
  & {\color{green!50!black} +4.10} & {\color{green!50!black} +2.60} & {\color{green!50!black} +3.30}
  & {\color{green!50!black} +4.20} & {\color{green!50!black} +2.40} & {\color{green!50!black} +4.80}
  & {\color{green!50!black} +4.60} & {\color{green!50!black} +1.90} & {\color{green!50!black} +1.80}
  & {\color{green!50!black} +2.10} \\

  Qwen2.5-VL-3B-it (FT)
  & {\color{red!60!black} -8.68} & {\color{red!60!black} -2.67} & {\color{red!60!black} -3.72}
  & {\color{green!50!black} +5.94} & {\color{green!50!black} +0.46} & {\color{red!60!black} -2.67}
  & {\color{green!50!black} +7.25} & {\color{green!50!black} +2.29} & {\color{green!50!black} +8.56}
  & {\color{green!50!black} +7.78} & {\color{green!50!black} +6.73} & {\color{green!50!black} +4.37}
  & {\color{green!50!black} +2.14} \\

  Qwen2.5-VL-7B-it (FT)
  & {\color{red!60!black} -12.99} & {\color{red!60!black} -0.99} & {\color{red!60!black} -6.46}
  & {\color{green!50!black} +3.19} & {\color{red!60!black} -2.29} & {\color{red!60!black} -3.07}
  & {\color{green!50!black} +9.72} & {\color{green!50!black} +3.71} & {\color{green!50!black} +7.89}
  & {\color{green!50!black} +9.98} & {\color{green!50!black} +1.62} & {\color{green!50!black} +7.37}
  & {\color{green!50!black} +1.48} \\

  Qwen3-VL-4B-it (FT)
  & {\color{red!60!black} -1.22} & {\color{red!60!black} -2.26} & {\color{red!60!black} -0.43}
  & {\color{green!50!black} +5.31} & {\color{green!50!black} +3.22} & {\color{green!50!black} +3.23}
  & {\color{green!50!black} +9.75} & {\color{green!50!black} +5.31} & {\color{green!50!black} +7.40}
  & {\color{green!50!black} +8.70} & {\color{green!50!black} +6.88} & {\color{green!50!black} +6.10}
  & {\color{green!50!black} +4.33} \\

  Qwen3-VL-8B-it (FT)
  & {\color{green!50!black} +1.07} & {\color{red!60!black} -1.03} & {\color{red!60!black} -1.81}
  & {\color{green!50!black} +7.60} & {\color{green!50!black} +7.85} & {\color{green!50!black} +8.63}
  & {\color{green!50!black} +10.46} & {\color{green!50!black} +8.89} & {\color{green!50!black} +12.03}
  & {\color{green!50!black} +13.08} & {\color{green!50!black} +5.24} & {\color{green!50!black} +11.77}
  & {\color{green!50!black} +6.98} \\

  Qwen3-VL-32B-it (FT)
  & {\color{green!50!black} +0.03} & {\color{red!60!black} -1.67} & {\color{red!60!black} -3.25}
  & {\color{green!50!black} +6.26} & {\color{green!50!black} +7.11} & {\color{green!50!black} +7.59}
  & {\color{green!50!black} +9.42} & {\color{green!50!black} +7.60} & {\color{green!50!black} +11.24}
  & {\color{green!50!black} +12.04} & {\color{green!50!black} +3.85} & {\color{green!50!black} +11.08}
  & {\color{green!50!black} +5.94} \\

  \bottomrule
  \end{tabular}
  \caption{Relative improvement from supervised fine-tuning on CinematicVQA, reported as the change $\Delta$ (percentage points) over each model's Zero-Shot baseline.
  {\color{green!50!black}Green}/{\color{red!60!black}red} denote gains/losses.}\label{tab:finetune}
  \end{center}
  \vspace{-0.15in}
\end{table*}

Overall, \cqat{} fine-tuning yields a clear and consistent improvement over the corresponding vanilla models. For instance, SFT increases the average accuracy from 42.87 to 47.20 for Qwen3-VL-4B-it and from 42.83 to 49.81 for Qwen3-VL-8B-it. The gains are not uniform across question types: SFT fine-tuning primarily strengthens higher-level semantic reasoning—most notably \textit{Narrative Function} and \textit{Multi-Hop}, while improvements in \textit{Technique Recognition} are limited and can even slightly regress, suggesting that \cqat{} fine-tuning more strongly improves the model’s ability to \emph{explain} and \emph{interpret} cinematic effects than to reliably discriminate fine-grained, taxonomy-specific technique labels. This may partly reflect that our fine-tuning is not specifically designed for fine-grained technique recognition. Meanwhile, the detailed expert narrations provide stronger supervision for interpreting visual presentation and narrative effects. Finally, we observe a scale effect among the fine-tuned variants: larger backbones tend to achieve stronger overall performance and exhibit larger gains on semantics-heavy questions, indicating better capacity to absorb the structured supervisory signal and generalize to unseen videos.

\section{Conclusion and Future Work}

In this work, we introduce \cqa{}, the first benchmark for comprehensive cinematic reasoning in LVLMs. Rather than evaluating isolated technique recognition, \cqa{} examines whether models can connect cinematographic choices to visual presentation and narrative intent. To support this, we propose the Cinematic Scene Graph (CSG), which represents each clip through structured technique--presentation--narrative relations. Experiments on \cqae{} show that current LVLMs can often describe visual patterns but struggle to identify the underlying film techniques, suggesting weak film-grammar priors; Chain-of-Thought prompting provides limited benefit. Fine-tuning on \cqat{} consistently improves higher-level interpretation, including narrative function and multi-hop reasoning, while technique recognition remains challenging.

We believe \cqa{} is a starting point for film-literate video understanding. Several directions remain promising for future work. First, our current focus on \emph{Composition}, \emph{Lighting}, and \emph{Camera Operation} reflects their foundational role in cinematography and the limited availability of large-scale, high-quality annotations for cinematic understanding. While \cqa{} currently relies on expert annotations from VADB~\cite{qiao2025vadb}, future work can construct CSGs from broader data sources, such as large-scale YouTube UGC videos and professionally captured cinematic datasets, together with larger-scale human subjective studies to develop a more comprehensive cinematic taxonomy. Second, the structured nature of CSG provides a natural foundation for reinforcement learning and reward design, supporting process supervision along cinematic dependencies such as $technique \rightarrow presentation \rightarrow narrative$. Third, future benchmarks can extend beyond cinematographic techniques to incorporate semantic understanding, including events, characters, and scene context, as well as multimodal elements such as audio, dialogue, and script metadata. In particular, soundtracks and sound effects are important for conveying scene atmosphere and narrative intent but are not explicitly modeled in the current \cqa{} framework. Finally, CSG-based representations may also support media generation as an interpretable evaluation framework or reward signal for producing videos with coherent cinematic style and narrative intent.

\bibliographystyle{IEEEtran}
\bibliography{strings,refs}

\end{document}